\documentclass[11pt]{article}
\usepackage{acl2017}
\usepackage{times}
\usepackage{latexsym}
\usepackage{color}
\usepackage{colortbl}
\usepackage [english]{babel}
\usepackage [autostyle, english = american]{csquotes}
\MakeOuterQuote{"}
\usepackage[colorlinks]{}
\usepackage{pgfplots,wrapfig}
\usepackage{tikz}
\usepackage{subcaption}

\usepackage{color}
\usepackage{booktabs}
\usepackage{url}
\newcommand{\ra}[1]{\renewcommand{\arraystretch}{#1}}
\newcolumntype{P}[1]{>{\centering\arraybackslash}p{#1}}
\pgfplotsset{compat=1.12}
\usepackage{array, caption, floatrow, tabularx, makecell, booktabs}%
\usepackage{bold-extra}

\usepackage{tikz}
\usetikzlibrary{positioning,calc}

\tikzset{
  redondo/.style={
    draw=blue,
    line width=1pt,
    rounded corners=3pt,
    text width=#1
  },
  punto/.style={
    fill=red,
    circle,
    inner sep=1.25pt
  },
  tresp/.pic={
    \node[punto] at (0.25,0) {};
    \node[punto] at (0.5,0) {};
    \node[punto] at (0.75,0) {};
  },
  dosp/.pic={
    \node[punto] at (0.25,0) {};
    \node[punto] at (0.5,0) {};
  },
  cuadra/.style={
    fill=teal,
    minimum size=10pt
  },
  arr/.style={
    line width=1pt,
    draw=green!70!black,
    ->,
    >=latex
  }  
}

\aclfinalcopy 

\title{Are you serious?: Rhetorical Questions and Sarcasm \\ in Social Media Dialog}

   \author{
   \textbf{Shereen Oraby$^1$, Vrindavan Harrison$^1$, Amita Misra$^1$,} 
   \textbf{Ellen Riloff $^2$ and Marilyn Walker$^1$} \\
   $^1$ University of California, Santa Cruz \\
   $^2$ University of Utah \\ 
    {\tt \{soraby,vharriso,amisra2,mawalker\}@ucsc.edu} \\
   {\tt riloff@cs.utah.edu}  \\ 
   }

\date{}

\begin{document}

\maketitle

\begin{abstract}
Effective models of social dialog must understand a broad range of
rhetorical and figurative devices. Rhetorical questions (\textbf{RQs})
are a type of figurative language whose aim is to achieve a pragmatic
goal, such as structuring an argument, being persuasive, emphasizing a point, 
or being ironic. While there are computational models for other forms of
figurative language, rhetorical questions have received little
attention to date.  We expand a small dataset from previous work, presenting a corpus of 10,270 RQs from debate
forums and Twitter that represent different discourse functions. We show that we can clearly distinguish between RQs and
sincere questions (0.76 F1). We then show that RQs can be used both sarcastically and non-sarcastically, observing that non-sarcastic (other) uses of RQs are frequently argumentative in forums, and persuasive in tweets.
We present experiments to distinguish between these uses of RQs using SVM and LSTM models that
represent linguistic features and post-level context, achieving results
as high as 0.76 F1 for {\sc sarcastic} and 0.77 F1 for {\sc other} in forums, and 0.83 F1 for both {\sc sarcastic} 
and {\sc other} in tweets. We supplement our quantitative experiments with an in-depth
characterization of the linguistic variation in RQs.
\end{abstract}


\section{Introduction}

Theoretical frameworks for figurative language posit eight standard
forms: {\it indirect questions, idiom, irony and sarcasm, metaphor, simile,
hyperbole, understatement,} and {\it rhetorical questions}
\cite{RobertsKreuz94}.  While computational models have been developed
for many of these forms, rhetorical questions (\textbf{RQs}) have
received little attention to date.  Table \ref{tbl:rqs} shows examples of RQs from social media in debate forums
and Twitter, where their use is prevalent.

\begin{table}
\begin{small}
\begin{subtable}[t!]{1.0\linewidth}
\ra{1.3}
\begin{tabular}
{@{}p{0.1cm}|p{7cm}@{}}
\toprule
1 & {\bf Then why do you call a politician who ran such measures liberal} {\it OH yes, it's because you're a
 republican and you're not conservative at all.} \\ \hline
2 & {\bf Can you read?} {\it You're the type that just waits to say
your next piece and never attempts to listen to others.} \\ \bottomrule
   \noalign{\vskip 0.05in}  
3 & {\bf Pray tell, where would I find the atheist church?  }{\it
Ridiculous.} \\ \hline
4 & {\bf You lost this debate Skeptic, why drag it back
 up again?} {\it There are plenty of other subjects
 that we could debate instead.} \\
\bottomrule               
 \end{tabular}
 \caption{RQs in Forums Dialog}\label{tbl:forums-rqs}

\vspace{0.3cm}

\begin{tabular}
{@{}p{0.1cm}|p{7cm}@{}}
\toprule
5 & {\bf Are you completely revolting?} {\it Then you should slide into my DMs, because apparently thatÕs the place to be. \#Sarcasm}\\ \bottomrule
   \noalign{\vskip 0.05in}  
6 & {\bf Do you have problems falling asleep?} {\it Reduce anxiety, calm the mind, sleep better naturally [link]}\\ \hline
7 & {\bf The officials messed something up?} {\it  I'm shocked I tell you.SHOCKED.} \\ \hline
8 & {\bf Does ANY review get better than this?} {\it From a journalist in New York.} \\
\bottomrule               
 \end{tabular}
  \caption{RQs in Twitter Dialog\\}\label{tbl:twitter-rqs}
\end{subtable}
\end{small}
 \caption{RQs and Following Statements in Forums and Twitter Dialog}\label{tbl:rqs}
\end{table}

RQs are defined as utterances
that have the structure of a question, but which are {\it not
  intended} to seek information or elicit an answer
\cite{Rohde06,Frank90,ilie1994else,Sadock71}. RQs are often used in
arguments and expressions of opinion, advertisements and other
persuasive domains \cite{Pettyetal81RQ}, and are frequent in social
media and other types of informal language.  

Corpus creation and
computational models for some forms of figurative language have been
facilitated by the use of hashtags in Twitter, e.g. the {\tt
  \#sarcasm} hashtag
\cite{BammanSmith15,Riloffetal13,Liebrechtetal13}. Other figurative
forms, such as similes, can be identified  via
lexico-syntactic patterns \cite{Qadiretal16,Qadiretal15,VealeHao07}. RQs are not
marked by a hashtag, and their syntactic form is indistinguishable
from standard questions \cite{Han02,Sadock71}.

Previous theoretical work examines the
discourse functions of RQs and compares the overlap in discourse
functions across all forms of figurative language
\cite{RobertsKreuz94}. For RQs, 72\% of subjects assign {\it to clarify} as a
function, 39\% assign {\it discourse management}, 28\% mention {\it to
  emphasize}, 56\% percent of subjects assign negative emotion, and
another 28\% mention positive emotion.\footnote{Subjects could provide
  multiple discourse functions for RQs, thus the
  frequencies do not add to 1.}  The discourse functions of clarification, discourse
management and emphasis  are
clearly related to argumentation.  One of the other largest overlaps in
discourse function between RQs and other figurative forms is between
RQs and irony/sarcasm (62\% overlap), and there are many studies
describing how RQs are used sarcastically
\cite{Gibbs00,ilie1994else}. 

To better understand the relationship between RQs and irony/sarcasm,
we expand on a small existing
dataset of RQs in debate forums from our previous work \cite{Orabyetal16}, ending up
with a corpus of 2,496 RQs and the self-answers or statements that follow them. We use the heuristic described in that work to collect a completely novel corpus of 7,774 RQs from Twitter. Examples from our final dataset of 10,270 RQs and their following self-answers/statements are shown in Table \ref{tbl:rqs}. We observe great diversity in the use of
RQs, ranging from sarcastic and mocking (such as the forum post in Row 2), 
to offering advice based on some anticipated answer (such as the tweet in Row 6).

In this study, we first show that RQs can clearly be distinguished from sincere, information-seeking
questions (0.76 F1). Because we are interested in how RQs are used sarcastically, we define our task as distinguishing sarcastic uses from other uses RQs, observing that non-sarcastic RQs are often used argumentatively in forums (as opposed to the more mocking sarcastic uses), and persuasively in Twitter (as frequent advertisements and calls-to-action). To distinguish between sarcastic and other uses, we perform classification experiments using SVM and LSTM models, exploring different levels of context, and showing that adding linguistic features improves classification
  results in both domains.

This paper provides the first in-depth investigation of the use of RQs in
different forms of social media dialog. We present a novel task,
dataset\footnote{The Sarcasm RQ corpus will be available at: \\ {\url {https://nlds.soe.ucsc.edu/sarcasm-rq}}.}, and results aimed at understanding how RQs can be recognized,
and how sarcastic and other uses of RQs can be distinguished. 

\section{Related Work}
\label{related-work-sec}

Much of the previous work on RQs has focused on
RQs as a form of figurative language, and on describing their
discourse functions \cite{schaffer-rqs,Gibbs00,RobertsKreuz94,Frank90,Pettyetal81RQ}.
Related work in linguistics has primarily focused on the differences
between RQs and standard questions \cite{Han02,ilie1994else,Han1a}. For
example Sadock \shortcite{Sadock71} shows that RQs can be followed by
a {\it yet} clause, and that the discourse cue {\it after all} at the
beginning of the question leads to its interpretation as an RQ.
Phrases such as {\it by any chance} are primarily used on information
seeking questions, while negative polarity items such as {\it lift a
finger} or {\it budge an inch} can only be used with RQs, e.g. {\it Did John
help with the party?} vs. {\it Did John lift a finger to help with the party?}

RQs were introduced into the DAMSL coding scheme when it was
applied to the Switchboard corpus
\cite{SWBD-DAMSL}. To our knowledge, the only computational work
utilizing that data is by Battasali et al. \shortcite{Bhattasalietal15}, who used n-gram language
models with pre- and post-context to distinguish RQs from regular
questions in SWBD-DAMSL. Using context improved their results to 0.83 F1 on a
balanced dataset of 958 instances, demonstrating that context information
could be very useful for this task. 

Although it has been observed in the literature that RQs are often used sarcastically \cite{Gibbs00,ilie1994else},
previous work on sarcasm classification has not focused on RQs
\cite{BammanSmith15,Riloffetal13,Liebrechtetal13,Filatova12,gonzalezetal11,Davidovetal10,Tsuretal10}.
\newcite{Riloffetal13} investigated the utility of sequential features
in tweets, emphasizing a subtype of sarcasm that consists of an
expression of positive emotion contrasted with a negative situation,
and showed that sequential features performed much better than
features that did not capture sequential information. More recent work on sarcasm 
has focused specifically on sarcasm identification on Twitter using neural network approaches 
\cite{Poria16,Ghosh16,Zhang16,Amir16}.

Other work emphasizes features of semantic incongruity in recognizing
sarcasm \cite{Joshietal15,Reyesetal12}.  Sarcastic RQs clearly feature
semantic incongruity, in some cases by expressing the certainty of
particular facts in the frame of a question, and in other cases by
asking questions like {\it "Can you read?"} (Row 2 in
Table~\ref{tbl:rqs}), a competence which a speaker must
have, prima facie, to participate in online discussion.

To our knowledge, our previous work is the first to consider the task of distinguishing sarcastic vs. not-sarcastic RQs, where we construct a corpus of sarcasm in three types: generic, RQ, and hyperbole, and provide simple baseline experiments using ngrams (0.70 F1 for {\sc sarc} and 0.71 F1 for {\sc not-sarc}) \cite{Orabyetal16}. Here, we adopt
the same heuristic for gathering RQs and expand the corpus in debate forums, also collecting a novel Twitter corpus. 
We show that we can distinguish between {\sc sarcastic} and {\sc other} uses of RQs that we observe, such as argumentation and persuasion in forums and Twitter, respectively. We show that linguistic features aid in the classification task, and explore the effects of context, using traditional and neural models.


\section{Corpus Creation}
\label{data-sec}

Sarcasm is a prevalent discourse function of RQs. In previous work, we observe both sarcastic and not-sarcastic uses of RQs in forums, and collect a set of sarcastic and not-sarcastic RQs in debate by using a heuristic stating that an RQ is a question that occurs in the middle of a turn, and which is answered immediately by the speaker themselves \cite{Orabyetal16}. RQs are thus defined {\it intentionally}: the speaker indicates that their intention is not to elicit an answer by not ceding the turn.\footnote{We acknowledge that this method may miss RQs that do not follow this heuristic, but opt to use this conservative pattern for expanding the data to avoid introducing extra noise.}

In this work, we are interested in doing a closer analysis of RQs in social media. We use the same RQ-collection heuristic from previous work to expand our corpus of {\sc sarcastic} vs. {\sc other} uses RQs in debate forums, and create another completely novel corpus of RQs in Twitter. We observe that the other uses of RQs in forums are often argumentative, aimed at structuring an argument more emphatically, clearly, or concisely, whereas in Twitter they are frequently persuasive in nature, aimed at advertising or grabbing attention. Table \ref{tbl:rqs-sarc-notsarc} shows examples of 
sarcastic and other uses of RQs in our corpus, and we describe our
data collection methods for both domains below.\\

\begin{table}
\begin{small}
\begin{subtable}[h]{1.0\linewidth}
\ra{1.3}
\begin{tabular}
{@{}p{0.1cm}|p{7cm}@{}}
\multicolumn{2}{c}{\bf \sc Sarcastic} \\ \toprule
      1 & {\bf Do you even read what anyone posts?} {\it Try it, you
might learn something.......maybe not.......} \\\hline
2   &  {\bf If they haven't been discovered yet,
    HOW THE BLOODY HELL DO YOU KNOW?} 
    {\it Ten percent more brains and you'd be pondlife.} \\ \bottomrule
   \noalign{\vskip 0.1in}  
   
\multicolumn{2}{c}{\bf \sc Other} \\ \toprule
 3  &  {\bf How is that related to deterrence? }{\it Once again,
    deterrence is preventing through the fear of consequences.} \\ \hline
  4 &  {\bf Well, you didn't have my experiences, now did you? }{\it  Each woman who has an abortion could have
    innumerous circumstances and experiences.} \\
\bottomrule               
 \end{tabular}
 \caption{{\sc Sarc} vs. {\sc Other} RQs in Forums}\label{tbl:forums-sarc-notsarc}

\vspace{0.5cm}

\begin{tabular}
{@{}p{0.1cm}|p{7cm}@{}}
\multicolumn{2}{c}{\bf \sc Sarcastic} \\ \toprule
      5 & {\bf When something goes wrong, what's the easiest thing to do?} {\it Blame the victim!
      Obviously they had it coming \#sarcasm \#itsajoke \#dontlynchme} \\\hline
6   &  {\bf You know what's the best?} 
    {\it Unreliable friends. They're so much un. \#sarcasm \#whatever.} \\ \bottomrule
   \noalign{\vskip 0.1in}  
   
\multicolumn{2}{c}{\bf \sc Other} \\ \toprule
 7  &  {\bf And what, Socrates, is the food of the soul? }{\it Surely,
 I said, knowledge is the food of the soul. Plato} \\ \hline
  8 &  {\bf Craft ladies, salon owners, party planners?} {\it You need to state your \#business [link]}\\ 
\bottomrule               
 \end{tabular}
   \caption{{\sc Sarc} vs. {\sc Other} RQs in Twitter\\}\label{tbl:twitter-sarc-notsarc}
\end{subtable}
\end{small}
 \caption{Sarcastic vs. Other Uses of RQs}\label{tbl:rqs-sarc-notsarc}
\end{table}

\noindent{\bf Debate Forums:} 
The Internet Argument Corpus (IAC 2.0) \cite{Abbottetal16}
contains a large number of  discussions about politics and social issues, making it
a good source of RQs.  Following our previous work \shortcite{Orabyetal16}, we first extract RQs in posts
whose length varies from 10-150 words, and collect five annotations for each of 
the RQs paired with the context of
their following statements. 

We ask Turkers to specify whether or not
the RQ-response pair is sarcastic, as a binary question.  We count a post as "sarcastic" if
the majority of annotators (at least 3 of the 5) labeled
the post as sarcastic. Including the 851 posts per class from previous work \cite{Orabyetal16},
this resulted in 1,248 sarcastic posts out of 4,840 (25.8\%), a
significantly larger percentage than the estimated 12\% sarcasm ratio in
debate forums \cite{Swansonetal14}.  We then balance the 1,248
sarcastic RQs with an equal number of RQs that 0 or 1 annotators
voted as sarcastic, giving us a total of 2,496 RQ pairs. For our experiments, 
all annotators had above 80\% agreement with the majority vote. \\

\noindent{\bf Twitter:}  We also extract RQs defined as above from a
set of 80,000 tweets with a \texttt{\#sarcasm}, \texttt{\#sarcastic},
or \texttt{\#sarcastictweet} hashtag. We use the hashtags as "labels",
as in other work \cite{Riloffetal13,Reyesetal12}.  This yields 3,887 sarcastic RQ tweets, again
balanced with 3,887 RQ pairs from a set of random tweets (not
containing any sarcasm-related hashtags). We remove all sarcasm-related hashtags and
username mentions (prefixed with an "@") from the posts, for a total of
7,774 total RQ tweets. 


\section{Experimental Results}

In this section, we present experiments classifying rhetorical vs. information-seeking
questions, then sarcastic vs. other uses of RQs.

\subsection{RQs vs. Information-Seeking Qs}
\label{rhet-fact-sec}

By definition, fact-seeking questions are not RQs. We take advantage of
the annotations provided for subsets of the IAC, in particular the
subcorpus that distinguishes {\sc factual} posts from {\sc emotional} posts \cite{Abbottetal16,Orabyetal15}.\footnote{\url{https://nlds.soe.ucsc.edu/factfeel}}
  Table \ref{table:examples-fact-rq} shows examples of {\sc factual/info-seeking} questions.
  
\begin{table}[h]
\begin{small}
\centering
\ra{1.3}
\caption{Examples of Information-Seeking Questions}
\begin{tabular}
{@{}p{0.1cm}|p{7cm}@{}}
\multicolumn{2}{c}{\bf \sc Factual/Info-Seeking Questions} \\ \toprule
      1 & {How do you justify claims about  covering only a fraction more ?} \\ \hline
      2 & {If someone is an attorney or in law enforcement, would you please give an interpretation?} \\ \bottomrule         
 \end{tabular}
 \label{table:examples-fact-rq}
\end{small}
\end{table}

To test whether {\sc RQ} and {\sc factual/info-seeking} questions are easily distinguishable, we randomly select a sample of 1,020 questions from our forums RQ corpus, and balance them with the same number of questions from {\sc fact} corpus. We divide the question data into 80\% train and 20\% test, and use an
SVM classifier \cite{Pedregosaetal11}, with GoogleNews Word2Vec (W2V) \cite{Mikolovetal13} features. We perform a grid-search on our training set using 3-fold cross-validation for parameter tuning, and report results
on our test set. Table \ref{table:fact-rq-results} shows the precision (P), recall (R) and F1 scores we achieve, showing good classification performance for distinguishing both classes, at 0.76 F1 for the {\sc RQ} class, and 0.74 F1 for the {\sc factual/info-seeking} class. 

\begin{table}[h]
\begin{small}
\centering
\ra{1.3}
\caption{ \label{table:fact-rq-results}Supervised Learning Results for RQs vs. Fact/Info-Seeking Questions in Debate Forums}
\begin{tabular}
{@{}p{0.1cm}|P{2cm}|P{1cm}|P{1cm}|P{1cm}@{}}\toprule
\bf \# & \bf Class &  \bf P  &  \bf R  &  \bf F1 \\ \midrule
1 & {\sc RQ} & 0.74 & 0.79 & 0.76 \\
2 & {\sc Fact} & 0.77 & 0.72 & 0.74 \\
\bottomrule               
 \end{tabular}
\end{small}
\end{table}

\subsection{Sarcastic vs. Other Uses of RQs}
\label{sarc-notsarc-sec}
\label{ml-res}

Next, we focus on distinguishing {\sc sarcastic} from {\sc other} uses of RQs in forums and Twitter.  We divide the full RQ data from each domain (2,496 forums and 7,774 tweets, balanced between the two classes) into 80\%
train and 20\% test data. We experiment with two models, an SVM classifier from Scikit Learn \cite{Pedregosaetal11}, and a bidirectional LSTM model \cite{chollet2015keras} with a TensorFlow backend \cite{Abadi16}. We perform a grid-search using cross-validation on our training set for parameter tuning, and report results on our test set.

For each of the models, we establish a baseline with W2V features (Google News-trained Word2Vec size 300 \cite{Mikolovetal13} for the debate forums, and Twitter-trained Word2Vec size 400
\cite{godinetal15}, for the tweets). We experiment with different embedding representations, finding that we achieve best results by averaging the word embeddings for each input when using SVM, and creating an embedding matrix (number of words by embedding size for each input) as input to an embedding layer when using LSTM.\footnote{In future work, we plan to further explore the effects of different embedding representations on model performance.} 

  \begin{figure}[t!]
\begin{center}
   \includegraphics[width=2.5in]{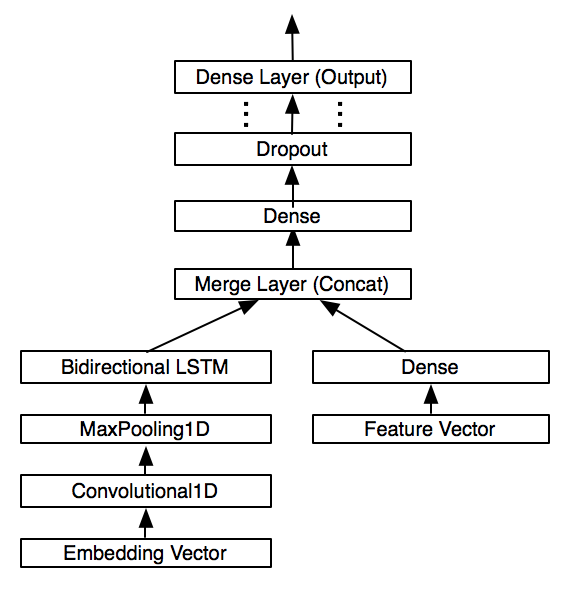}
  \caption{LSTM Network Architecture}
  \label{network-fig}
  \end{center}
\end{figure}

For our LSTM model, we experiment with various different layer architectures from previous work \cite{Poria16,Ghosh16,Zhang16,Amir16}. For our final model
(shown in Figure \ref{network-fig}), we use a sequential
embedding layer, 1D convolutional layer, max-pooling, a bidirectional LSTM, dropout
layer, and a sequence of dense and dropout layers with a final sigmoid activation layer for the
output.  

For additional features, we experiment with using post-level scores (frequency of each category in the input, normalized by word count) from the Linguistic Inquiry and Word Count (LIWC) tool \cite{PennebakerFrancis01}. We experiment
with which LIWC categories to include as features on our training data, and end up with a set of 20 categories for each domain\footnote{We
discuss some of the highly-informative LIWC categories by domain in Sec.~\ref{analysis-sec}.}, as shown in Table \ref{table:liwc-features}. When adding features to the LSTM model, we include a dense and merge layer to concatenate features, followed by the dense and dropout layers and sigmoid output. 

\begin{table}[h]
\begin{small}
\centering
\ra{1.3}
\caption{LIWC Features by Domain}
\begin{tabular}
{@{}p{3.5cm}|p{3.5cm}@{}}
\bf Debate Forums & \bf Tweets \\ \toprule
\sc $2^{nd}$ Person & \sc $2^{nd}$ Person \\ 
\sc $3^{rd}$ Person Plural & \sc $3^{rd}$ Person Plural \\
\sc $3^{rd}$ Person Singular & \sc Articles \\
\sc Adverbs & \sc Auxiliary Verbs \\
\sc Affiliation & \sc Certainty \\
\sc Assent & \sc Colon \\
\sc Auxiliary Verbs & \sc Comma \\
\sc Compare & \sc Conjunction \\
\sc Exclamation Marks &\sc  Friends \\
\sc Focus Future &\sc  Male \\
\sc Friends & \sc Negations \\
\sc Function & \sc Negative Emotion \\
\sc Health & \sc Parenthesis \\
\sc Informal & \sc Quote Marks \\
\sc Interrogatives & \sc Risk \\
\sc Netspeak & \sc Sadness \\
\sc Numerals & \sc Semicolon \\
\sc Quantifiers & \sc Swear Words \\
\sc Rewards & \sc Word Count \\
\sc Sadness & \sc  Words per Sentence \\
 \bottomrule         
 \end{tabular}
 \label{table:liwc-features}
\end{small}
\end{table}
    
We experiment with different levels of textual context in training for both the forums
and Twitter data (keeping our test set constant, always testing on only the RQ and self-answer portion of the text). We are motivated by the intuition that training on larger context will help us identify more informative segments of RQs in test. Specifically, we test four different levels of context representation:     
    \begin{itemize}  \setlength\itemsep{-0.5em}
    \item {$RQ$}: only the RQ and its self-answer
    \item {$Pre + RQ$}: the preceding context and the $RQ$
    \item {$RQ + Post$}: the $RQ$ and following context
    \item {$ Full Text$}: the full text or tweet (all context)
    \end{itemize}

Table \ref{table:results} presents our results on the classification
task by model for each domain, showing P, R, and F1 scores for each class (forums in Table \ref{table:forums-results} and Twitter in Table \ref{table:tweet-results}). For each domain, we present the same experiments for both models (SVM and LSTM), first showing a W2V baseline (Rows 1 and 6 in both tables), then adding in LIWC (Rows 2 and 7), and finally presenting results for W2V and LIWC features on different context levels (Rows 2-5 for SVM and Rows 7-10 for LSTM). \\

\begin{table*}[t]
\begin{small}
\begin{subtable}[h]{1.0\linewidth}
\ra{1.3}
\begin{tabular}
{@{}p{0.2cm}|p{0.9cm}|p{0.7cm}|p{2.8cm}|p{1.7cm}||P{0.9cm}|P{0.9cm}|P{0.9cm}||P{0.9cm}|P{0.9cm}|P{0.9cm}@{}}\toprule
\multicolumn{5}{c||}{} & \multicolumn{3}{c||}{\bf \sc Sarcastic} & \multicolumn{3}{c}{  \sc Other} \\ \midrule
\bf \#  & \bf Domain & \bf Model & \bf Features  & \bf Training &  \bf P  &  \bf R  &  \bf F1 & \bf P & \bf R  & \bf F1 \\  
\noalign{\vskip 0.05in}\midrule
1 & \bf Forums  & SVM  & $W2V_{Google}$  & $RQ$ & 0.74 & 0.70 & 0.72 & 0.71 & 0.75 & 0.73  \\ \hline
2 & &   &                           $W2V_{Google} + LIWC$ & $RQ$ & 0.78 & 0.74 & \bf 0.76 & 0.75 & 0.79 & \bf 0.77 \\
3 & &               &                & $Pre + RQ$ & 0.76 & 0.72 & 0.74 & 0.73 & 0.78 & 0.76 \\
4 & &               &                & $RQ + Post$ & 0.75 & 0.76 & 0.75 & 0.76 & 0.74 & 0.75 \\
5 & &               &                & $Full$ $Text$ & 0.75 & 0.77 &\bf 0.76 & 0.76 & 0.74 & 0.75  \\ \hline

 6   & &  LSTM &                           $W2V_{Google}$ & $RQ$ & 0.76 & 0.62 & 0.68 & 0.68 & 0.80 & 0.74 \\\hline

7 & &   &                           $W2V_{Google} + LIWC$ & $RQ$ & 0.76 & 0.68 & 0.72 & 0.71 & 0.79 & 0.75 \\
8 & &               &                & $Pre + RQ$ & 0.81 & 0.60 & 0.69 & 0.68 & 0.86 & 0.76 \\
9 & &               &                & $RQ + Post$ & 0.74 & 0.76 & 0.75 & 0.76 & 0.74 & 0.75 \\
10 & &               &                & $Full$ $Text$ & 0.76 & 0.67 & 0.71 & 0.70 & 0.78 & 0.74  \\
\bottomrule               
 \end{tabular}
 \caption{Supervised Learning Results on Debate Forums}\label{table:forums-results}

\vspace{0.5cm}

\begin{tabular}
{@{}p{0.2cm}|p{0.9cm}|p{0.7cm}|p{2.8cm}|p{1.7cm}||P{0.9cm}|P{0.9cm}|P{0.9cm}||P{0.9cm}|P{0.9cm}|P{0.9cm}@{}}\toprule
\multicolumn{5}{c||}{} & \multicolumn{3}{c||}{\bf \sc Sarcastic} & \multicolumn{3}{c}{  \sc Other} \\ \midrule
\bf \#  & \bf Domain & \bf Model & \bf Features  & \bf Training &  \bf P  &  \bf R  &  \bf F1 & \bf P & \bf R  & \bf F1 \\  
\noalign{\vskip 0.05in}\midrule
1 & \bf Twitter & SVM & $W2V_{Tweet}$  & $RQ$ & 0.77 & 0.85 & 0.80 & 0.83 & 0.74 & 0.78 \\ \hline
2 & &   &                           $W2V_{Tweet} + LIWC$ & $RQ$ & 0.80 & 0.86 & \bf 0.83 & 0.85 & 0.79 & 0.82 \\
3 & &               &                & $Pre + RQ$ & 0.80 & 0.87 & \bf 0.83 & 0.86 & 0.78 & 0.82 \\
4 & &               &                & $RQ + Post$ & 0.79 & 0.87  & \bf 0.83 & 0.86 & 0.77 & 0.81 \\
5 & &               &                & $Full$ $Text$ & 0.80 & 0.86 & \bf 0.83 & 0.85 & 0.79 & 0.82 \\ \hline

6    &  &    LSTM     &                   $W2V_{Tweet}$  &$RQ$ & 0.76 & 0.70 & 0.73 & 0.72 & 0.78 & 0.75   \\\hline
7  & &         &                   $W2V_{Tweet} + LIWC$ & $RQ$ & 0.80 & 0.82 & 0.81 & 0.82 & 0.79 & 0.80 \\
8  & &       &                       & $Pre + RQ$ & 0.78 & 0.84 & 0.81 & 0.83 & 0.76 & 0.80\\
9  & &       &                       & $RQ + Post$ & 0.83 & 0.81 & 0.82 & 0.82 & 0.84 & \bf 0.83 \\
10 & &       &                      & $Full$ $Tweet$ & 0.80 & 0.83 & 0.82 & 0.83 & 0.79 & 0.81 \\
\bottomrule                
 \end{tabular}
 \caption{Supervised Learning Results on Twitter}\label{table:tweet-results}
\end{subtable}
\end{small}
 \caption{Supervised Learning Results for RQs in Debate Forums and Twitter}\label{table:results}
\end{table*}

\noindent {\bf Debate Forums}: 
From Table \ref{table:forums-results}, for both models, we observe that the addition of LIWC features gives us a large improvement over the baseline of just W2V features, particularly for the {\sc sarc} class (from 0.72 F1 to 0.76 F1 {\sc sarc} and 0.73 F1 to 0.77 F1 {\sc other} for SVM in Rows 1-2, and from 0.68 F1 to 0.72 F1 {\sc sarc} and 0.74 F1 to 0.75 F1 {\sc other} for LSTM in Rows 6-7). Our best results come from the SVM model, with best scores of 0.76 F1 for {\sc sarc} and 0.77 {\sc other} in Row 2 from using only the RQ and self-response in training (with the same F1 for {\sc sarc} when training on the full text). 

We observe that while the SVM results with LIWC features do not change significantly depending on the training context (Rows 3-5), the LSTM model is highly sensitive to context changes for the {\sc sarc} class (Rows 8-10).
Some interesting findings emerge when training on different context granularities for LSTM: 
  our best LSTM results for the {\sc sarc} class come from training on the $RQ+Post$ context 
  (0.75 F1 in Row 9), and for the $Pre+RQ$ context for the {\sc other} class (0.76 F1 in Row 8). We note that this increase in the {\sc sarc} class from plain word
embeddings to word embeddings combined with LIWC and context
is larger than the increase in the {\sc other} class, indicating that post-level context for 
{\sc sarc} captures more diverse instances in training. We also note that these results beat our previous baselines using only
  ngram features on the smaller original dataset of 851 posts per class (0.70 F1 for {\sc sarc}, 0.71 F1 for {\sc not-sarc}) \cite{Orabyetal16}.

We investigate why certain context features benefit each class differently for LSTM. Table~\ref{table:context}
shows examples of single posts, divided into $Pre$, $RQ$, and $Post$. Looking at Row 1,
it is clear that while the RQ and self-answer portion may
not appear to be sarcastic, the $Post$ context makes the sarcasm
much more pronounced. This is frequent in the case of sarcastic debate posts, where
the speaker often ends with a sharp remark or an interjection (like {\it "gasp!!!"}), or emoticons (like winking {\it ;)} 
or roll-eyes {\it 8-)}). In the case of the {\sc other} forums posts, the RQ is often
nestled within sequences of questions, or other RQ and self-answer pairs (Row 2).

\begin{table}[h]
\begin{footnotesize}
\begin{subtable}[t]{1.0\linewidth}
\ra{1.3}
\begin{tabular}
{@{}p{0.1cm}|p{0.8cm}|p{5.8cm}@{}}
\multicolumn{3}{c}{\bf \sc Sarcastic} \\ \toprule
      1 &  {\sc $Pre$} & [...] the argument I hear most often from so-called
  'pro-choicers' is that you cannot legislate morality. \\ \cline{2-3}
  & {\sc $RQ$} & {\bf Well then
    what can you legislate? }{\it Every law in existence is legislation of
    morality!} \\\cline{2-3}
    & {\sc $Post$}&  By that way of thinking, then we should have no
  laws. If someone kidnaps and murders your 3-year-old child, then
  let's hope the murderer goes free because we cannot legislate
  morality! \\ \bottomrule
   \noalign{\vskip 0.1in}  
\multicolumn{3}{c}{\bf \sc Other} \\ \toprule
 2  &   {\sc $Pre$} & what that man did isn't illegal in the us? you couldn't claim self defence if someone running away like that.   \\\cline{2-3}
&   {\sc $RQ$} & {\bf you think that the fact that man had a gun stopped people getting shot? }{\it what would have happened if he hadn't would be that the robbers got away with some money. }\\\cline{2-3}
  &  {\sc $Post$} & nothing to do with taking lives. [...]\\ \bottomrule                 
 \end{tabular}
 \caption{{\sc Sarc} vs. {\sc Other} RQs in Context on Forums}\label{tbl:forums-sarc-notsarc}
\vspace{0.3cm}
\begin{tabular}
{@{}p{0.1cm}|p{1cm}|p{5.8cm}@{}}
\multicolumn{3}{c}{\bf \sc Sarcastic} \\ \toprule
 3  &   {\sc $Pre$} & Gasp! \\\cline{2-3}
 & {\sc $RQ$} & {\bf Two football players got into it with each other?! }{\it How uncivilized! }\\\cline{2-3}
 &   {\sc $Post$} &    Lets make a big deal about it! \#NFLlogic \#cowboys\\ \bottomrule    
   \noalign{\vskip 0.1in}  
\multicolumn{3}{c}{\bf \sc Other} \\ \toprule
 4  &   {\sc $Pre$} &\\\cline{2-3}
 & {\sc $RQ$} & {\bf Are you willing to succeed? }{\it The answer isn't as simple as you may think.}\\\cline{2-3}
 &   {\sc $Post$} & Read my blog post and you'll see why.... [link] \\ \bottomrule               
 \end{tabular}
   \caption{{\sc Sarc} vs. {\sc Other} RQs in Context on Twitter\\}\label{tbl:twitter-sarc-notsarc}
\end{subtable}
\end{footnotesize}
 \caption{Sarcastic vs. Other Uses RQs in Context}
 \label{table:context}
\end{table}


\noindent {\bf Twitter:} From Table \ref{table:tweet-results}, we observe that the best result of 0.83 F1 for the {\sc sarc} class come from the SVM model (for all context levels), while the best result of 0.83 F1 for the {\sc other} class comes from the LSTM model.  We observe a strong performance increase from adding in LIWC features for both models, even more pronounced than for forums (0.80 F1 to 0.83 F1 {\sc sarc} and 0.78 F1 to 0.82  F1 {\sc other} for SVM in Rows 1-2, and 0.73 F1 to 0.81 F1 {\sc sarc} and 0.75 F1 to 0.80 F1 {\sc other} for LSTM in Rows 6-7). 

Again, while the SVM results do not vary based on changes in context, there is a large improvement in the {\sc other} class for LSTM when using $RQ+Post$ level context, giving us our best {\sc other} class results. From Table \ref{table:liwc-tweets} Row 4, we see an example of a "call-to-action" that are frequent and distinctive in non-sarcastic Twitter
RQs, asking users to visit a link at the end of a tweet ($Post$ RQ). In the case of the {\sc sarc} tweet in Row 3, 
the extra tweet-level context (such as initial exclamations/interjections) aids in highlighting the sarcasm, but is limited in length compared to the forums posts, explaining the smaller gain from context in the Twitter domain for {\sc sarc}. 

\label{ling-res}
\begin{table*}[t!]
\begin{footnotesize}
  \begin{floatrow}[2]
    \ttabbox%
   {\begin{tabularx}
    {0.48\textwidth}{p{0.1cm}|P{0.5cm}|p{1.4cm}|p{4cm}}
        \multicolumn{4}{c}{\bf \sc Sarcastic} \\ \toprule
      \# & \bf FW & \bf Feature &\bf Example \\ \hline  \noalign{\vskip 0.1in}  
1 & 15.19 & $2^{nd}$ Person & {\bf Do you ever read headers?} {\it You got a mouth on you as big as grand canyon.} \\\hline
2 & 12.09 & Informal & {\bf The hate you're spewing is palpable, yet you can't even see that can you?} {\it Hypocrites, ya gotta luv em.} \\\hline
3 & 8.92 & Exclamation & {\bf Force the children to learn science?}  {\it How obscene!!} \\\hline
4 & 4.66 & Netspeak & {\bf To make fun of my title?} {\it lol, how that stings...}  \\   \noalign{\vskip 0.15in}    
 
         \multicolumn{4}{c}{\bf \sc Other} \\ \toprule
               \# & \bf FW & \bf Feature &\bf Example \\ \hline  \noalign{\vskip 0.1in}  
5 & 8.98 & Interrog. & {\bf How do you know it's the truth?} {\it If it were definitive [...]}  \\\hline
6 & 8.54 & $3^{rd}$ Person Plural & {\bf what's the difference?} {\it both are imposing their ideologies} \\\hline
7 & 3.93 & {Quantifiers} & {\bf [...] we have minimum wage, why can't we have a maximum wage?} {\it some of  [...]}  \\ \hline
8 & 3.88 & Health & {\bf When will the people press congress to take up abortion?} {\it It's the job of congress [...]} \\
      \bottomrule
      \end{tabularx}}
    {\caption{Forums LIWC Categories}
      \label{table:liwc-forums}}
    \hfill%
    \ttabbox%
    {\begin{tabularx}
    {0.48\textwidth}{p{0.1cm}|P{0.5cm}|p{1.4cm}|p{4cm}}
      
         \multicolumn{4}{c}{ \bf  \sc Sarcastic} \\ \toprule
                \# & \bf FW & \bf Feature &\bf Example \\ \hline \noalign{\vskip 0.1in}  
1 & 15.71 & {Comma} & {\bf Wait, wait, I can't...it's impossible...NO WAY?!} {\it - a stiffer track pad?! } \\ \hline       
2 & 6.86 & Word Count & {\bf Shouldn't you be in power? }{\it You know best after all. } \\ \hline         
3 & 5.89 & Negations & {\bf Can't we do that already without brain imaging?} {\it I think it's called empathy} \\ \hline
4 & 3.91 & $3^{rd}$ Person Plural & {\bf How intelligent, they make the laws and then violate [them]?} {\it That is absurd!} \\  \noalign{\vskip 0.15in}  
       \multicolumn{4}{c}{\bf \sc Other} \\ \toprule
             \# & \bf FW & \bf Feature &\bf Example \\ \hline  \noalign{\vskip 0.1in}  
5 & 4.51 & Swear Words & {\bf Idk why I'm fighting my sleep?!} {\it Ain't shit else to do} \\ \hline
6 & 3.60 & Risk & {\bf Have their been launch pad explosions? } {\it That would be a risk. } \\ \hline
7 & 3.01 & $2^{nd}$ Person & {\bf Do you want a great deal on [...]? } {\it Check out the latest } \\ \hline
8 & 2.83 & Friends & {\bf Can I get 12.7k followers today? } {\it :) xo Thanks to everyone who is following me.} \\
      \bottomrule
      \end{tabularx}}
    {\caption{Tweet LIWC Categories}
      \label{table:liwc-tweets}}
  \end{floatrow}
  \end{footnotesize}
\end{table*}%

Comparing both domains, we observe that the results for tweets in Table \ref{table:tweet-results} are much higher than the results for forums in Table \ref{table:forums-results}, noting that this could be a result of less lexical diversity and a larger amount of data, making them more distinguishable than the more varied forums posts. We plan to explore these differences more extensively in future work.


\section{Linguistic Characteristics of RQs by Class and Domain}
\label{analysis-sec}
In this section, we discuss linguistic characteristics we observe in our {\sc sarcastic} vs {\sc other} uses of RQs using the most informative LIWC features.

Previous work has observed that {\sc factual} utterances
  are often very heavy on technical jargon \cite{Orabyetal15}: this is
also true of factual questions. When analyzing differences in LIWC categories in our factual vs. RQ data, we find that our factual questions are slightly longer on average than the RQs (14 words on average compared to 12). We also find
significant differences in "function" word categories ($p<0.05$, unpaired t-test) in LIWC, marking use of personal references, and "affective processes" ($p<0.005$). Both categories are more
  prevalent in the {\sc RQs} than in the {\sc fact} questions, indicating
  more emotional language that is targeted towards the second party.

A qualitative analysis of our {\sc sarcastic} vs. {\sc other} data shows that sarcastic RQs
in forums are often followed by short statements that serve to point
attention or mock, whereas the other RQ-self-response pairs often serve as a
technique to concisely structure an argument.  RQs in
Twitter are frequently advertisements (persuasive
communication) \cite{Pettyetal81RQ}, making them more distinguishable from the more diverse
sarcastic instances. Tables
\ref{table:liwc-forums} and \ref{table:liwc-tweets} show examples of
LIWC features that are most characteristic of each domain
  and class based on our experiments. For ranking, we show the learned feature weight 
  (FW) for each class, found by performing 10-fold 
  cross-validation on each training set using an SVM model with only LIWC features.

In Table \ref{table:liwc-forums}, Row 1, we observe that $2^{nd}$
person mentions are frequent in the sarcastic debate forums posts
(referring to the other person in the debate), while in the Twitter
domain, they come up as significant features in the
\textit{non-sarcastic} tweets, where they are used as methods to
persuade readers to interact: click a link, like, comment, share (Table
\ref{table:liwc-tweets}, Row 6). Likewise, "informal" words and more
"verbal speech style" non-fluencies, including exclamations and social
media slang ("netspeak"), also appear in sarcastic debate (Table
\ref{table:liwc-forums}, Rows 2 and 4). Features of sarcastic forums
include exclamations (Table \ref{table:liwc-forums}, Rows 3), often used in a hyperbolic
or figurative manner \cite{McCarthy2004,RobertsKreuz94}. We find that sarcastic tweets
frequently include sets of exclamations/interjections strung together with commas (Table \ref{table:liwc-tweets}, Row 1), and are often shorter than the tweets in the non-sarcastic class (Table \ref{table:liwc-tweets}, Row 3).

Table \ref{table:liwc-forums} shows that "interrogatives" are a
strong feature of argumentative forums (Row 7), as well as the use of technical jargon (including quantifiers health words with some domain-specific topics, such as abortion) 
(Row 8). Table \ref{table:liwc-tweets} indicates that {\sc other} 
tweets frequently contain forms of advertisement and calls-to-action
involving $2^{nd}$ person references (Row 7). Similarly, RQ tweets are sometimes used to express frustration ("swear words" in Row 5), or increase engagement with references to "friends" and followers (Row 8).


\section{Conclusions}
\label{conc-sec}
In this study, we expand on a small corpus from previous work to create a large corpus
of RQs in two domains where RQs are prevalent: debate forums and Twitter. To our knowledge, this is the first in-depth study dedicated to sarcasm and other uses of RQs in social media. We present supervised learning experiments using traditional and neural models to classify sarcasm in each domain, providing analysis of unique features across domains and classes, and exploring the effects of training of different levels of context.

We first show that we can distinguish between information-seeking and rhetorical questions (0.76 F1). We then focus on classifying sarcasm in only the RQs, showing that there are distinct linguistic differences
between the methods of expression used in RQs across forums and Twitter. For forums, we show
that we are able to distinguish between the sarcastic and other uses (noting they are often argumentative) in forums with 0.76 F1 for {\sc sarc} and 0.77 F1 for {\sc not-sarc}, improving on our baselines from previous work on a smaller dataset \cite{Orabyetal16}.

We also explore sarcastic and other uses of RQs on Twitter, noting that other non-sarcastic uses of RQs are often advertisements, a form of persuasive communication not represented
in debate dialog. We show that we can distinguish between sarcastic
and other uses of RQ in Twitter with scores of 0.83 F1 for both the {\sc sarc} and {\sc other} classes. We observe
that tweets are generally more easily distinguished than the more diverse forums, and that the addition of linguistic categories from LIWC greatly improves classification performance. We also note that the LSTM model is more sensitive to context changes than the SVM model, and plan to explore the differences between the models in greater detail in future work.

Other future work also includes expanding our dataset to capture more
instances of what may characterize RQs across these
domains to improve performance, and also to analyze other interesting domains, such as Reddit. We believe that it will be possible to improve our results 
by using more robust models, and also by developing features to represent the  \textit{sequential}
properties of RQs by further utilizing the larger context of
the surrounding dialog in our analysis.

\section*{Acknowledgments}
\label{ack}
This work was funded by NSF CISE RI 1302668,
under the Robust Intelligence Program. 

\bibliography{nl}
\bibliographystyle{acl_natbib}

\end{document}